\documentclass[conference]{IEEEtran}
\IEEEoverridecommandlockouts

\usepackage[T1]{fontenc}
\usepackage[]{lmodern}
\usepackage[utf8]{inputenc}
\usepackage[english]{babel}

\usepackage{lipsum}
\usepackage{url}
\usepackage{xspace}
\usepackage[]{siunitx}
\usepackage[ruled,linesnumbered]{algorithm2e}
\usepackage{rotating}
\usepackage{subcaption}
\usepackage{multirow}
\usepackage{multicol}
\usepackage[bookmarks=false]{hyperref}
\hypersetup{
  colorlinks   = true, 
  urlcolor     = blue, 
  linkcolor    = blue, 
  citecolor   = red 
}

\usepackage{cite}
\usepackage{amsmath,amssymb,amsfonts}
\usepackage{algorithmic}
\usepackage{graphicx}
\usepackage{textcomp}
\usepackage[dvipsnames, table]{xcolor}
\def\BibTeX{{\rm B\kern-.05em{\sc i\kern-.025em b}\kern-.08em
    T\kern-.1667em\lower.7ex\hbox{E}\kern-.125emX}}

\newcommand{\murts}{\textsc{\si{\micro}RTS}\xspace}
\newcommand{\microrts}{\textsc{microRTS}\xspace}
\newcommand{\poadaptive}{\textsc{POAdaptive}\xspace}
\newcommand{\microphantom}{\textsc{microPhantom}\xspace}
\newcommand{\ie}{\textit{i.e.}}
\newcommand{\csp}{\textsc{CSP}\xspace}
\newcommand{\cp}{\textsc{CP}\xspace}
\newcommand{\cop}{\textsc{COP}\xspace}
\newcommand{\ghost}{\textsc{GHOST}\xspace}
\newcommand{\rdu}{\textsc{RDU}\xspace}

\begin{document}

\title{\microphantom: Playing \microrts\\under uncertainty and chaos}

\author{\IEEEauthorblockN{Florian Richoux}
 \IEEEauthorblockA{\textit{JFLI, CNRS, National Institute of Informatics / Université de Nantes}\\
Tokyo, Japan\\
florian.richoux@polytechnique.edu}
}

\maketitle

\begin{abstract}
  This  competition  paper  presents   \microphantom,  a  bot  playing
  \microrts   and    participating   in   the   2020    \microrts   AI
  competition. \microphantom is based on our previous bot \poadaptive
  which  won the  partially  observable  track of  the  2018 and  2019
  \microrts   AI   competitions.   In   this   paper,   we  focus   on
  decision-making under  uncertainty, by tackling the  Unit Production
  Problem  with  a  method  based   on  a  combination  of  Constraint
  Programming and decision  theory.  We show that using  our method to
  decide  which units  to train  improves significantly  the win  rate
  against the second-best \microrts  bot from the partially observable
  track.   We  also show  that  our  method  is resilient  in  chaotic
  environments, with a  very small loss of efficiency  only.  To allow
  replicability and to facilitate further research, the source code of
  \microphantom is  available, as  well as the  Constraint Programming
  toolkit it uses.
\end{abstract}


\section{Introduction}\label{sec:introduction}

Recently, the  Game AI  community has  seen a  strong increase  in the
number  of  available  AI  competitions  and  environments.   Although
competitions  can  be great  tools  to  stimulate and  accelerate  the
research in  Game AI,  they may  also bring  a major  drawback: having
scripted, hard-coded bots  tailored to win a  competition, rather than
taking  risks by  creating new  AI techniques  and improving  existing
ones.

\microrts is  a minimalist real-time  strategy game developed to  be a
convenient  environment  to  test  and  improve  Game  AI  techniques,
historically  Monte  Carlo  Tree   Search  techniques  to  tackle  the
combinatorial  multi-armed bandit  problem~\cite{microRTS}. Like  more
complex  game environments  such as  StarCraft, \microrts  contains an
imperfect  information environment  with a  fog of  war masking  enemy
units   and   buildings.   What   \microrts   offers   besides  is   a
non-deterministic  environment  and  requires  less  engineering  than
StarCraft  to  make   a  bot.  Thanks  to   partially  observable  and
non-determinism tracks, \microrts  AI competitions propose challenging
environments  that  push  participants  to  go  beyond  a  simple  but
efficient scripted bot.

In this paper,  we present \microphantom, our new  \microrts bot based
on \poadaptive. The  later was our \microrts bot  that participated in
the  2018  and 2019  \microrts  AI  competitions. The  decision-making
methods  used  in  \poadaptive  has been  described  in  our  previous
paper~\cite{AR19}  published  in  CEC  2019  proceedings.   Therefore,
Section~\ref{sec:poadaptive}  briefly  introducing  \poadaptive  shows
nothing     new,     except     the     competition     results     in
Section~\ref{sec:poadaptive_competitions}.

Like \poadaptive,  \microphantom focuses on a  decision-making problem
under uncertainty, the Unit Production Problem, implemented and solved
in Constraint Programming within our \ghost toolkit\footnote{Available
  at
  \href{https://github.com/richoux/GHOST}{github.com/richoux/GHOST}}~\cite{RUB16}.
This  is where  the name  \microphantom comes  from.  \poadaptive  and
\microphantom are developed in Java,  like \microrts, but \ghost being
a C++  toolkit, the decision-making  problem is  coded in C++  and the
constraint solver executable is called within the Java code.

This paper is organized as follows: Section~\ref{sec:microrts} gives a
short   presentation   of   \microrts   and   its   AI   competitions.
Section~\ref{sec:decision_making}   introduce   our  Unit   Production
Problem  as well  as  Constraint Programming  and  the Rank  Dependent
Utility, necessary to understand how  decision making works within our
bots.  In Section~\ref{sec:poadaptive}, we summarize a presentation of
\poadaptive    's   decision-making    method   from    our   previous
paper~\cite{AR19},   and   Section~\ref{sec:microphantom}   introduces
\microphantom and what  is new compared to  \poadaptive.  This section
contains  an  analysis  of  experimental  results  to  attest  to  the
efficiency  of  our  decision-making method  in  partially  observable
environments.  Finally, the paper concludes with some perspectives.

\section{\microrts}\label{sec:microrts}

In this section, we briefly present  the game \microrts and its annual
AI competition. 

\subsection{The game}\label{sec:microrts_game}

\microrts, or \murts, is an open-source, minimalist real-time strategy
(RTS)   game    developed   by    Santiago   Ontañón    for   research
purposes~\cite{microRTS}.

\begin{figure}
  \centering
  \includegraphics[width=0.8\linewidth]{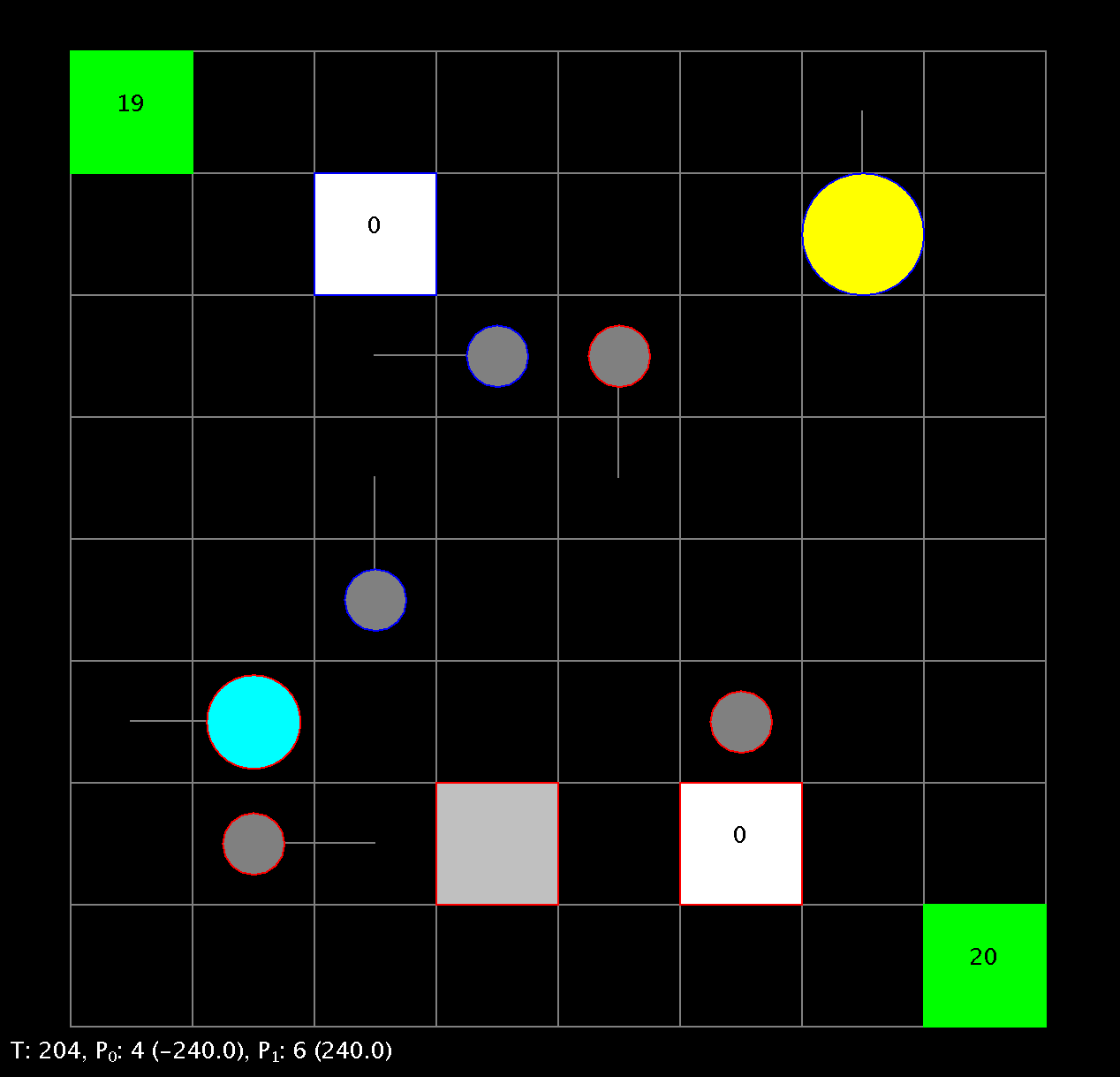}
  \caption{A game  frame of \microrts on  a 8x8 map. Resources  are in
    green, squares are buildings and round items are units.}
  \label{fig:microrts}
\end{figure}

\microrts provide to  players and researchers the  main mechanisms one
can find in  RTS games. The game  is played on a map,  here a discrete
grid.   Usually,  a  map  contains  several  resource  patches.   Once
collected, this  resource allows players  to make buildings  and train
units.   In  \microrts,  there  are  two  kinds  of  buildings:  bases
producing worker  units and stocking resources,  and barracks training
military units.  Four types of units are available: workers, and three
military  units: light,  ranged, and  heavy units.   Workers are  weak
against all units  but are the only ones able  to gather resources and
construct buildings.   All units  must be  on an  adjacent case  of an
enemy unit  or building to attack  it, except the ranged  unit able to
attack  at distance.   Units have  different attributes  such as  unit
costs, unit hit points, unit speed,  the time required to move, train,
harvest, etc.

A game is  played between two players, in 1v1.  The game is real-time,
meaning that players are doing their moves simultaneously, rather than
turn by turn as in most strategy games.  To win, a player must destroy
all enemy units and buildings.  If  no player reaches that goal before
a fixed number of game ticks, the game ends in a draw.

\microrts  supports  complete  and partially  observable  games,  \ie,
without or  with a fog  of war,  respectively, hiding enemy  units and
building  if they  are not  within the  sight range  of our  units and
buildings. \microrts also supports deterministic and non-deterministic
games, \ie,  where unit damages  are either deterministic  or randomly
drawn  within  a  fixed  range.    This  paper  focuses  on  partially
observable deterministic games.

\subsection{The competitions}\label{sec:microrts_competitions}

The first  \microrts competition  has been organized  in 2017  and was
hosted  by the  IEEE  Computational Intelligence  in Games~(CIG)  2017
conference~\cite{microRTScompetition}. Since 2017, an annual \microrts
competition is organized  at CIG, and now at CoG  since the conference
changed its name in 2019.

The three competitions from 2017 till 2019 were divided into 3 tracks:
the  standard track  (complete information  deterministic games),  the
non-deterministic   track   (complete  information   non-deterministic
games),  and  the  partially observable  track  (partially  observable
deterministic games).

Our bot \poadaptive  participated in the 2018 and  2019 competition in
the partially  observable track (also  in the standard track  in 2018,
even if the bot was designed to  deal with the fog of war). Results of
our   bots  in   the   partially  observable   track   are  given   in
Section~\ref{sec:poadaptive_competitions}.

The 2020 competition  will be hosted by CoG 2020  and will be composed
of 2 tracks only: the  classic track (previously named standard track)
and   the   partial  observability   track.    All   games  are   thus
deterministic. Our new  bot \microphantom will compete  in the partial
observability track; \poadaptive being removed from competitions.

In each track of each competition,  two different rankings of bots are
published: the ranking with open  maps, \ie, maps that were officially
listed  in the  rules of  the competition,  and the  ranking of  games
played  on  both  open  and   hidden  maps,  \ie,  unknown  maps  from
competitors.

\section{Decision-making under uncertainty}\label{sec:decision_making}

RTS  games   are  excellent   environments  to  develop   and  improve
decision-making  methods under  uncertainty.  Indeed,  such games  are
rich  enough  to contain  both  challenging  short-term and  long-term
decision-making  problems. Besides,  the  fog of  war implies  partial
observability of  the game state,  so players must take  both tactical
and strategic decisions under uncertainty.

\microphantom  is  part   of  a  research  project   aiming  to  solve
combinatorial   optimization   problems   under   uncertainty.    Many
decision-making   problems   can   be   expressed   as   combinatorial
optimization problems, as soon as one aims to optimize one value while
respecting  some  rules  or impossibilities,  giving  a  combinatorial
flavor to the  tackled problem.  Thus, in  particular, many RTS-related
decision-making   problems   can   be   expressed   as   combinatorial
optimization problems~\cite{RUB16}.

With \microphantom, we focus on the Unit Production Problem introduced
below. 

\subsection{The Unit Production Problem}\label{sec:problem}

As is usually the case in RTS games, units of \microrts follow a
rock-paper-scissors pattern.   Simulations introduced in  our previous
paper on the  topic show that heavy units are  efficient against light
units, which  are efficient against  ranged units, which are  in their
turn efficient against heavy units~\cite{AR19}.

The  question  captured by  the  Unit  Production Problem  is  simple:
without perfect information of the game state, in particular about the
enemy army  composition, and knowing  my army composition,  what units
should I produce to counter the enemy army?

This decision-making  problem under  uncertainty can  be modeled  as a
combinatorial  optimization  problem: we  need  to  decide what  units
should  we produce  next, such  that these  units integrated  into our
current army offer the best  counter to the partially-known enemy army
while verifying some constraints such as not producing more units than
our resource stock allows.

There exist  different paradigms  to model  combinatorial optimization
problems.   In this  paper, we  model the  Unit Production  Problem in
Constraint Programming.

\subsection{Constraint Programming}\label{sec:cp}

The basic  idea behind  Constraint Programming (\cp)  is to  deal with
combinatorial problems by  splitting them up into  two distinct parts:
the first part is modeling your problem via one Constraint Programming
formalism.  This is  usually done by a human being  and this task must
be ideally  easy and intuitive.   The second part consists  in finding
one  or several  solutions based  on your  model.  This  is done  by a
solver,  \ie,  a  program  running without  any  human  interventions.
Ideally, all the intelligence must be  placed in this second part, and
this is the main reason why \cp is part of Artificial Intelligence.

Constraint Programming proposes many formalisms to model problems; the
two most  well-known are  Constraint Satisfaction Problems  (\csp) and
Constrained  Optimization Problems  (\cop).   The former  is to  model
decision problems, \ie, problems where the answer is either yes or no;
the latter to model optimization problems, where we aim to maximize or
minimize a value computed by an objective function.

Moreover, several  formalisms dealing  with uncertainty exist  in \cp:
Mixed \csp,  Probabilistic \csp, Stochastic \csp~\cite{SCSP},  etc. We
recommend surveys~\cite{VJ05, Hnich2011} on this topic to get familiar
with these formalisms.  

Unfortunately, no truly convenient formalism  has been proposed in \cp
to model  a decision-making  problem where  constraints are  known and
crisp but  where the  value to optimize  depends upon  some stochastic
variables.  In other  words, our choices and  possibilities are known,
but a third-party agent we can  only partially observe, such as a game
environment with  imperfect information,  has a significant  impact on
the quality of our decisions.

In our previous paper~\cite{AR19}, we  presented a trick to model such
decision-making  problems  under  uncertainty with  the  regular  \cop
formalism,  and  exploiting results  from  decision  theory to  handle
uncertainty in the objective function.   One of the main advantages of
this trick is that one can use  a regular \cp solver since the problem
has been  modeled within  a regular  formalism. No  need to  develop a
specific, ad-hoc solver able to handle uncertainty.

In this paper, we propose a different \cp model to correct some issues
in our previous model. Moreover,  we model the Unit Production Problem
where constraints are replaced by  error functions.  This allows us to
make very powerful  models: where \csp or \cop models  offer a network
of constraints,  \ie, a network  of predicates expressing  if variable
assignments  satisfy or  not  each constraint,  our  model contains  a
network of error functions  expressing if variable assignments satisfy
the constraints or, if not, how  close they are to satisfy them.  This
allows  expressing a  finer  structure about  the  problem: the  error
functions network is  an ordered structure over  invalid assignments a
solver  can exploit  efficiently  to improve  the  search.  The  major
drawback is  that such models are  harder to define because  it is not
always obvious to find good error functions.

We  consider  error  function  networks  as  defined  by  Richoux  and
Baffier~\cite{RB20}.  Formally, our error  function network is defined
by a tuple ($V$, $D$, $F$) such that:
\begin{itemize}
\item $V$ is a set of variables,
\item $D$ is a domain, \ie, a set of values for variables in $V$,
\item  $F$  is   a  set  of  error  functions   with  different  scopes
  $\{x_1, \ldots, x_n\} \subseteq V$.
\end{itemize}

Error        functions        in       $F$        are        functions
$f :  D^n \rightarrow \mathbb{R}^+$ with  $n$ being the arity  of $f$.
An assignment $a$,  \ie, a tuple of  $n$ values (one value  in $D$ for
each of the $n$  variables in the scope of $f$), is  valid if and only
if $f(a) =  0$ holds. All other strictly positive  outputs of $f$ lead
to  invalid  assignments.  These  positive  outputs  of $f$  are  then
interpreted  like preferences  over  invalid  assignments: the  closer
$f(a)$ is to 0, the closer $a$ is to be a valid assignment for $f$.

Before introducing our  \cp model used in  \microphantom and comparing
it with the one used in our  previous bot \poadaptive, we give a short
introduction  on  Rank Dependent  Utility,  the  result from  decision
theory allowing us to handle uncertainty.

\subsection{Rank Dependent Utility}\label{sec:rdu}

Since   decision  theory   is  already   described  in   our  previous
paper~\cite{AR19}, we will go straight to the point in this section by
explaining what Rank Dependent Utility (\rdu)  is and how we use it in
our \cp models.

Rank     Dependent     Utility      has     been     introduced     by
Quiggin~\cite{RDU_quiggin_82,quiggin1993generalized}.     Like   other
notions  in decision  theory such  as Expected  Utility, \rdu  aims to
define a preference for decisions by associating a probability to each
possible consequence  of each possible decision.   But unlike Expected
Utility, it allows modeling attraction or repulsion to risks through a
probability deformation function.  This  can help to modify on-the-fly
the behavior of an agent making a decision regarding its environment.

Let $l$ be a vector of  consequences of an action and their associated
outcome probability, such that $l = (x_1, p_1; \ldots; x_n, p_n)$ with
$x_i$ a consequence and $p_i$  the probability that the decision leads
to consequence $x_i$.  The Rank Dependent Utility is then the function
defined by Equation~\ref{eq:rdu}.

\begin{figure*}[ht]
  \footnotesize
  \begin{equation}\label{eq:rdu}
    \rdu(l) = u(x_1) + \big(u(x_2) - u(x_1)\big) \times \phi\left(\sum_{i=2}^np_i\right) + \big(u(x_3) -
    u(x_2)\big)  \times  \phi\left(\sum_{i=3}^np_i\right) +  \ldots  +
    \big(u(x_n) - u(x_{n-1})\big) \times \phi(p_n)
  \end{equation}
\end{figure*}

\noindent
In  Equation~\ref{eq:rdu},  $u(x)$  is  a utility  function  over  the
consequence  space, intuitively  giving a  score to  consequences, and
$\phi(p)$  an  increasing function  from  $[0,  1]$  to $[0,  1]$  and
interpreted  as  a  probability deformation  function.   The  function
$\phi(p)$  can  be anything,  as  soon  as  it  is monotone  and  both
equalities $\phi(0)=0$ and $\phi(1)=1$  hold.  Consequences in $l$ are
ordered  such  that  $\forall  x_i,  x_j$   with  $i  <  j$,  we  have
$u(x_i) \leq u(x_j)$.

The  intuition behind  Equation~\ref{eq:rdu}  is  the following:  with
probability $p=1$, by making the given  decision, you are sure to have
at  least the  score of  the worst  consequence $x_1$,  \ie, $u(x_1)$.
Then, with (deformed) probability $\phi(p_2+\ldots+p_n)$, you can have
the     score     $u(x_1)$     plus     the     gain     equals     to
$\big(u(x_2) - u(x_1)\big)$.  With probability $\phi(p_3+\ldots+p_n)$,
you can have an additional gain equals to $\big(u(x_3) - u(x_2)\big)$,
and   so   on   until   having    an   additional   gain   equals   to
$\big(u(x_n)  -  u(x_{n-1})\big)$  with probability  $\phi(p_n)$.  The
obtained value  depends on  the order,  or rank, of  the value  of the
utility function  applied to consequences, justifying  the name ``Rank
Dependent Utility''.

\begin{figure*}[ht]
  \begin{subfigure}{0.3\textwidth}
    \centering
    \includegraphics[width=\linewidth]{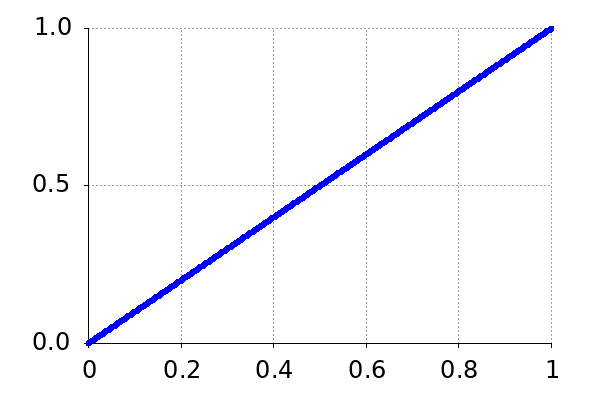}
    \caption{Neutral (identity)}\label{fig:neutral}
  \end{subfigure}
  \begin{subfigure}{0.3\textwidth}
    \centering
    \includegraphics[width=\linewidth]{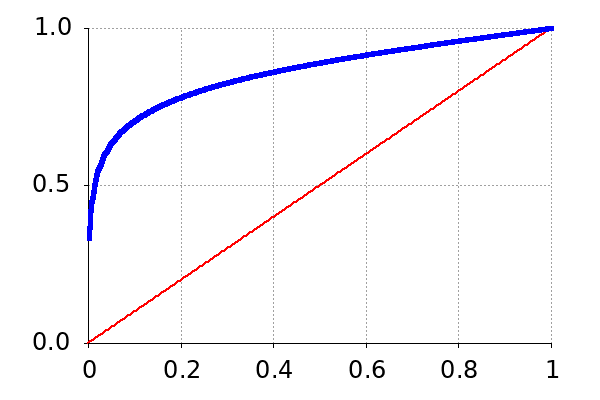}
    \caption{Optimistic (logit)}\label{fig:optimistic}
  \end{subfigure}
  \begin{subfigure}{0.3\textwidth}
    \centering
    \includegraphics[width=\linewidth]{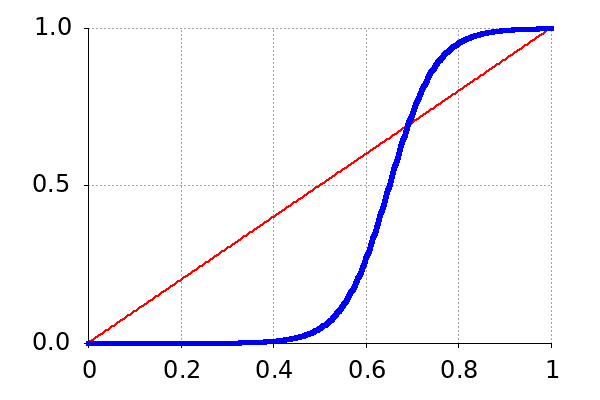}
    \caption{Pessimistic (logistic)}\label{fig:pessimistic}
  \end{subfigure}    
  \caption{Probability   deformation   functions    $\phi$   used   in
    \poadaptive and \microphantom.}
  \label{fig:phi_functions}
\end{figure*}

The   probability  deformation   function  $\phi$   allows  to   model
risk-aversion where a concave $\phi$ function defines an attraction to
risks and a convex $\phi$ function a repulsion to risks.  Intuitively,
if we  have $\phi(p)  \leq p$  for all  $p$, then  the agent  taking a
decision will underestimate  gains probabilities and then  will show a
kind of pessimism about risks.  We  will have the opposite behavior if
we have $\phi(p) \geq p$ for all  $p$. Instead of a convex function, a
sigmoid function can be used to model pessimism since it decreases the
probabilities  of good  outcomes  and increases  the probabilities  of
unfavorable  ones.    Figure~\ref{fig:phi_functions}  illustrates  the
identity, a logit  function, and a (shifted) logistic  function we use
in  our   bots  to   express  neutrality,  optimism,   and  pessimism,
respectively.

We have now everything we need  to present and compare \cp models used
in \poadaptive and \microphantom.

\section{\poadaptive: 2018 - 2019}\label{sec:poadaptive}

\poadaptive is  the name of  the bot  playing to \microrts  from which
\microphantom  is  based.   Its  source code  (CoG  2019  version)  is
available          on          our          GitHub          repository\footnote{
\href{https://github.com/richoux/microrts-uncertainty/releases/tag/cog2019}{github.com/richoux/microrts-uncertainty/releases/tag/cog2019}}. Its
main principles have  been detailed in our  CEC 2019 paper~\cite{AR19}
but we  recall here  its \cp model  and how  probability distributions
have been handled.

\subsection{\cp model}\label{sec:poadaptive_cp}

The  \cop  model  for  the  Unit  Production  Problem  implemented  in
\poadaptive  is fairly  simple, however  we  will only  give here  its
intuitive description.  The reader interested  in the formal  model is
invited to find it in~\cite{AR19}.

We need to  describe what are our decision variables  (the variable we
can  modify  the value),  the  stochastic  variables (handled  by  the
environment,  can   only  be  partially  observed),   domains  of  all
variables, our constraints, and our objective function to optimize.

Our model contains two kinds of {\bf decision variables}: 1. variables
$x_p$  to decide  the  number of  light, ranged,  and  heavy units  to
produce, and  2. variables $x_{ab}$ to  decide how many units  of type
$a$ must be ideally assigned to  fight against units of type $b$.  Our
{\bf  stochastic  variables}  $s_t$  represent the  number  of  light,
ranged, and heavy units composing the enemy army.

{\bf Domains}  are the same for  all variables: we consider  each unit
type  in the  game to  be between  0  and 20  units, which  is a  fair
assumption in \microrts.

{\bf Constraints} are crisps, meaning  there is no uncertainty within.
Therefore,  only decision  variables appear  in these  constraints. We
only have two  types of constraints: 1.  check if  the number of units
we aim  to produce  to not  exceed our  stock of  resource, and  2.  a
constraint   linking   variables   $x_p$  and   $x_{ab}$   such   that
$x_{al}  + x_{ah}  + x_{ar}  = x_a  + possess_a$  holds for  each type
$a \in \{l(ight), h(eavy), r(anged)\}$.  In other words, the number of
units of type $a$  we assigned to fight enemy units  must be equals to
the number of $a$ we plan to produce plus the number of $a$ we already
have.

Finally,    the   {\bf    objective   function}    is   to    maximize
$aim_l+aim_h+aim_r$ such that
\begin{displaymath}
  aim_b =  min\Big(1, \sum\limits_{a\in\{l,h,r\}}(x_{ab} \times  const_{ab}) -
  s_b\Big)
\end{displaymath}
where $const_{ab}$ is  a constant in $\mathbb{R}$  indicating how many
units of  type $a$ are  required to beat one  unit of type  $b$. These
constants have been  determined by running 200 skirmishes  of 10 units
against 10 units  of all possible combinations. The  minimum between 1
and the second part is here to forbid overkill, \ie, heavily defeating
one type of enemy unit at the expense of other types.

\subsection{\rdu}\label{sec:poadaptive_rdu}

Algorithm~\ref{algo:rdu} from~\cite{AR19} describes how to compute the
\rdu value  from the  objective function  $f$ of  our \cp  model.  The
principle is simple:  sample values of stochastic values  in the scope
of $f$ according to their  probability distribution (Line 3). Then use
$f$ as a  utility function to give  a score to the  decision, \ie, the
current  assignment of  the decision  variables (Line  4). Repeat  the
operation $k$ times, sort the $k$ outputs of $f$, and compute the \rdu
value according to Equation~\ref{eq:rdu} (Line 7).

\begin{algorithm*}[ht]
  \SetKwData{Left}{left}\SetKwData{This}{this}\SetKwData{Up}{up}
  \SetKwFunction{Union}{Union}\SetKwFunction{FindCompress}{FindCompress}
  \SetKwInOut{Input}{input}\SetKwInOut{Output}{output}

  \Input{A decision $d$,  \ie, a vector in $D^n$, with  $D$ the domain
    of decision variables $v_1, \ldots, v_n$}
  \Output{A preference on $d$, \ie, a real value (the higher the better)}
  \BlankLine
  Initialize an empty vector $x$ of size $k$, with $k$ a parameter for
  the number of wanted samples\;
  \For{$i=1$ \KwTo $k$}{
    Sample values for stochastic variables $s_1, \ldots, s_m$ according to their probability
    distribution\;
    \tcp{$f$  is  our objective  function,  taking  both decision  and
      stochastic variables}
    $x[i] \leftarrow f(v_1, \ldots, v_n, s_1, \ldots, s_m)$\;
  }
  Sort(x)\;
  \tcp{Considering each  sample has  a probability  $\frac{1}{k}$, computes
    RDU}
  RDU $\leftarrow x[1] + (x[2] - x[1]) \times \phi(\frac{k-1}{k}) + (x[3] -
  x[2]) \times \phi(\frac{k-2}{k}) + \ldots + (x[k] - x[k-1]) \times \phi(\frac{1}{k})$\;
  \Return{RDU}
  \BlankLine
  \caption{Estimating a preference on the decision $d$}
  \label{algo:rdu}
\end{algorithm*}

The  pessimistic function  we  use is  the  shifted logistic  function
$\phi(p) = \frac{1}{1 + exp( - \lambda (2 p - shift) )}$ where $p$
is the  probability and with parameters  $\lambda=10$ and $shift=1.3$.
The     optimistic     function      is     the     logit     function
$\phi(p)   =   1   +  \frac{log(   \frac{p}{2-p}   )}{\lambda}$   with
$\lambda=10$.      These      functions      are      depicted      in
Figure~\ref{fig:phi_functions}.

Observe that we consider a  uniform distribution among possible inputs
of the  objective function  $f$, \ie,  a probability  $\frac{1}{k}$ is
associated to each of the $k$ sampled inputs. We can do this since the
real  stochastic variables'  probability  distribution  is taken  into
account when  we draw values  of these stochastic  variables following
their probability distribution.

These  distributions are  made from  the  analysis of  800 replays  of
\microrts games from the 2017 competition. For each game tick and each
unit type, we counted unit occurrence.  These statistics are sharpened
by observations  while playing a  game: if  we observe for  instance 3
enemy light  units at the  same moment, we nullify  probabilities that
the  enemy has  0, 1  or  2 light  units  only, and  we normalize  the
remaining probabilities.

\subsection{Experimental results}\label{sec:poadaptive_xp}

To  evaluate our  decision-making process,  we run  games between  the
second-best bot of the competition, POLightRush bot, and four methods:
\poadaptive using \rdu  with a pessimistic $\phi$  function, \rdu with
an  optimistic  $\phi$ function,  \rdu  with  $\phi$ as  the  identity
function,  and finally  a baseline  bot  having the  same behavior  as
\poadaptive except  for the  unit production decision,  taken randomly
among the three military units.

We run  100 games  on each of  6 basic maps  of different  sizes, from
$8 \times 8$ to  $64 \times 64$ grids: 50 games  where our bot started
at the North-East  position, and 50 at the  South-West position. Then,
we compute the normalized score like in \microrts competitions: we sum
scores for each game, where the winner has a score of 1, the loser has
0, and both bots have 0.5 for  draw games, then we divide total scores
by the number of games played.

\begin{table}[ht!]
  \centering{
    \begin{tabular}{|c|c|c|c|}
      \hline
      Baseline & Neutral & Optimistic & Pessimistic\\
      \hline
      40.00 & 42.93 & 44.93 & 44.5\\
      \hline
    \end{tabular}
  }
  \caption{Score averages of 600 games (100 per map) against \poadaptive on basic maps}\label{tab:poadaptive_xp}
\end{table}

Table~\ref{tab:poadaptive_xp} compiles 2.400 games  in total and shows
averages of  normalized scores  for the  baseline bot  and \poadaptive
using a neutral, optimistic, and  pessimistic $\phi$ function.  We can
see that \poadaptive with an optimistic or pessimistic $\phi$ function
is  doing slightly  better than  the  baseline or  \poadaptive with  a
neutral probability deformation function.

\subsection{Competition results}\label{sec:poadaptive_competitions}

\poadaptive participated in the partially observable track of the 2018
and 2019  \microrts AI competitions. Seven  participants registered to
the 2018 edition,  and six to the 2019 edition.  In both competitions,
four baseline AIs were among the participants.

\begin{figure*}[ht]
  \begin{subfigure}{0.45\textwidth}
    \centering
    \includegraphics[width=\linewidth]{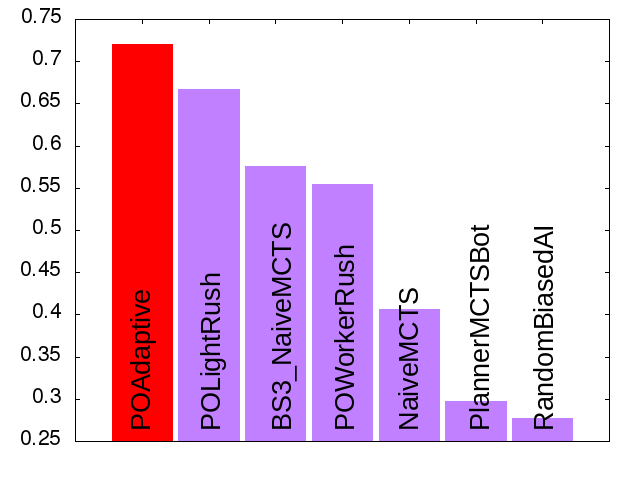}
    \caption{CIG 2018}\label{fig:cig2018}
  \end{subfigure}    
  \begin{subfigure}{0.45\textwidth}
    \centering
    \includegraphics[width=\linewidth]{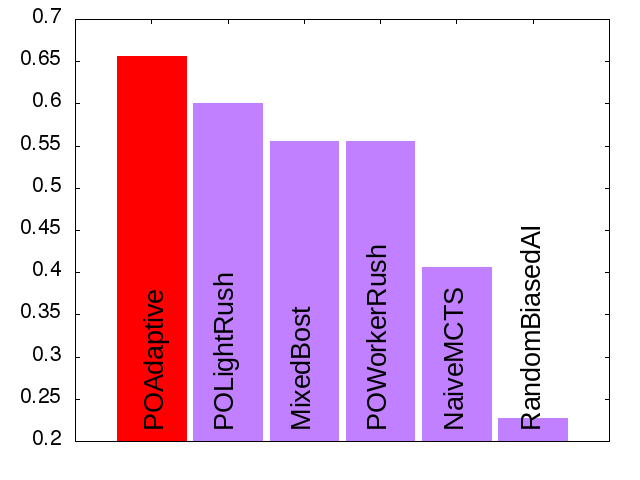}
    \caption{CoG 2019}\label{fig:cig2019}
  \end{subfigure}    
  \caption{Normalized  scores of the partially  observable track
    of the 2018 - 2019 \microrts competitions, on all maps.}
  \label{fig:competitions}
\end{figure*}

\poadaptive won  both the  2018 and  2019 partially  observable track,
both on open maps only and on all maps.  Figure~\ref{fig:competitions}
gives the final  normalized scores on all maps.  One  can see that the
POLightRush baseline bot was each  time the most challenging opponent,
this   is   why    experiments   in   Sections~\ref{sec:poadaptive_xp}
and~\ref{sec:microphantom_xp} take POLightRush as an opponent.

\section{\microphantom: 2020}\label{sec:microphantom}

\microphantom is  the new name  of \poadaptive, from which  we improve
the  \cp  model,  some  parts  of  the  code  to  be  robust  to  rule
modifications,  and  we  changed  the  way  to  make  the  probability
distribution of stochastic variables.  The source code, as well as all
experimental  setups   and  results,  are  available   on  our  GitHub
repository\footnote{\href{https://github.com/richoux/microPhantom/tree/develop}{github.com/richoux/microPhantom/tree/develop}}.

\subsection{Main differences with \poadaptive}\label{sec:microphantom_main}

\subsubsection{\cp model}

As  written in  Section~\ref{sec:cp}, we  propose this  time an  error
function-based model rather than a \cop model.

We  keep the  same decision  and stochastic  variables and  domains as
described  in   Section~\ref{sec:poadaptive_cp}.   We   defined  error
functions corresponding  to the  two types of  constraint in  the \cop
model, and add a third type of  error function to express the idea that
we cannot train more units than idle barracks we have. Our three kinds
of error function are then:
\begin{equation*}
  \begin{array}{ll}
    f_1: & max(0, stock - (x_l . cost_l + x_h . cost_h + x_r
           . cost_r)\\
    f_2: & abs(x_{al} + x_{ah} + x_{ar} - x_a - possess_a)\\
    f_3: & max(0, nb\_idle\_barracks - (x_l + x_h + x_r))
  \end{array}
\end{equation*}

We also  changed the objective  function to  penalize the fact  that a
guessed number  of enemy units  of a given type  is not beaten  by the
current assignment. We do this using the following function:

\begin{equation}
  reg(x) = \left\{
    \begin{array}{rl}
      -(x^2+1) &\text{if } x < 0,\\
      x &\text{otherwise}.
    \end{array}\right.
  \label{eq:reg}
\end{equation}

\begin{figure}[ht]
  \centering
  \includegraphics[width=0.7\linewidth]{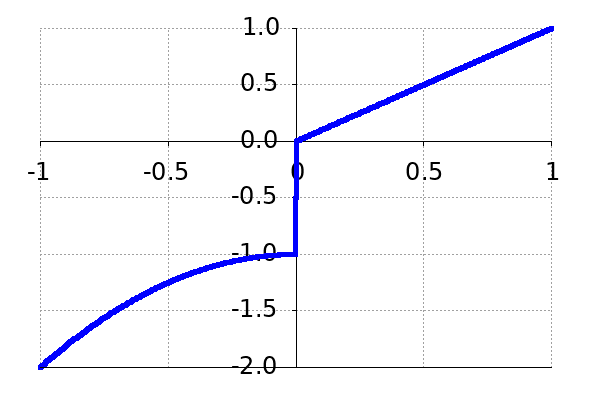}
  \caption{Our regulation function}\label{fig:reg}
\end{figure}

The function of Equation~\ref{eq:reg} is what we called a regulation function and
is illustrated in Figure~\ref{fig:reg}. 

Our       objective      function       becomes      \textit{maximize}
$reg(aim_l)  +  reg(aim_h)  +  reg(aim_r)$  with  $aim_a$  defined  in
Section~\ref{sec:poadaptive_cp}.   Injecting  this regulation  in  our
objective  function  gives it  a  new  interpretation while  using  an
optimistic or  pessimistic probability  deformation function:  with an
optimistic $\phi$  function, we  will favor  assignments that  fit the
best what  we think in  average what the  enemy composition is,  so we
maximize our  composition to be  the best counter. With  a pessimistic
$\phi$ function, we  will consider more worst cases for  us, \ie, that
the opponent is making a good counter against our army, so we maximize
resilience here, to be prepared for  the worst. In other words, we try
to minimize bad surprises.

\subsubsection{Stochastic variables estimation}

For  \poadaptive,  we  made   the  probability  distributions  of  our
stochastic variables  by analyzing replays,  and we draw the  value of
each stochastic variable following its  distribution and the number of
game ticks.   The issue with  that method  is that replays  we analyzed
won't certainly fit the strategy of our current opponent.

To get  around this  problem, we  now start each  game with  a uniform
distribution for a  unique random variable representing  the chance to
draw  one light,  heavy, or  ranged unit  from the  enemy army.   This
distribution is updated such that probability for each value (\ie, for
each type of units $t$), we have
\begin{equation*}
  \begin{array}{lll}
    \mathcal{P}(t) & = & \frac{1 + 2 \times obs(t) + past(t)}{total}\\
  \end{array}
\end{equation*}
where $obs(t)$ is the number of enemy  units of type $t$ we observe at
the  moment, $past(t)$  the  number  of enemy  units  of  type $t$  we
observed since  the beginning of  the game, minus destroyed  units and
units  in   $obs(t)$,  and   total  is  the   sum  of   numerators  of
$\mathcal{P}(l)$,  $\mathcal{P}(h)$,  and   $\mathcal{P}(r)$  to  have
$\mathcal{P}(l)+\mathcal{P}(h)+\mathcal{P}(r)=1$.    Enemy  units   we
currently observe count twice compared to  units we saw in the past to
be more reactive if the opponent  switches his or her strategy, and we
add $1$  on the numerator to  never have a probability  $0$ to produce
any unit type.

Then,  we  estimate  the  resource the  opponent  gathered  since  the
beginning of  the game, regarding  his or  her number of  workers, the
harvesting  time, worker's  speed,  how many  resources  a worker  can
carry, and the average distance of resource patches to the enemy base,
considering it  has a similar  placement as  our. We subtract  to this
estimation the sum of the cost of enemy units we destroyed and the sum
of the cost of enemy units we observed,  as well as the cost of a base
and  a barracks  if building  them was  required. What  remains is  an
estimation of the resource we do not know how it has been spent by the
opponent.

Now, we only draw  a value of our unique random  variable to know what
is the type of  an enemy unit we probably do not  see yet. We subtract
its  cost to  the  estimated  remaining enemy  resource  stock and  we
continue until not enough resource left. We add these estimated values
to the  number of enemy units  currently observed for each  unit type,
and we have  our estimation of the enemy army  composition.  We repeat
these draws each time we need to estimate the enemy army composition.

This way, we  have a sharper estimation of the  enemy army composition
which is  adapted to  the specific opponent  we are  currently playing
against.

\subsubsection{Chaotic-robust decision-making}

\microphantom  's  code contains  as  few  hard-coded game  values  as
possible.  Thus, any  attributes of the game can  change (cost of
units  and buildings,  time  to train/move/harvest,  hit points,  etc)
event during  the game,  and the decision-making  process will  not be
perturbed.   We call  \textit{chaotic environments}  such environments
where attributes are deterministic but change from game to game.

\subsubsection{In-game $\phi$ function replacement}

\poadaptive used an optimistic  or pessimistic probability deformation
$\phi$  function regarding  the  map  size, and  stick  with the  same
function during the whole game.

\microphantom can  switch from the  three $\phi$ functions  defined in
Section~\ref{sec:poadaptive_rdu} regarding  its current  situation. It
starts  with  a  neutral  $\phi$  function  and  will  switch  to  the
optimistic function if the sum of the cost of destroyed enemy units is
greater than the cost of its  destroyed units, added to twice the cost
of the  cheapest unit.  On  the other way  around, it switches  to the
pessimistic function.

\subsubsection{Domain knowledge-based actions}

\microphantom contains some hard-coded, domain knowledge-based actions
\poadaptive did  not have, such  as producing more workers  if several
resource patches are  near our base, and building more  barracks if it
is gathering resources faster than it can spend them.

\subsection{Experimental results}\label{sec:microphantom_xp}

Some domain knowledge-based actions  we added to \microphantom improve
the bot significantly. To measure improvements due to our non-scripted
modifications, namely the new \cp  model, the new stochastic variables
sampling and the in-game $\phi$  function replacement, we run the same
experiments than  in Section~\ref{sec:poadaptive_xp} on 6  basic maps,
with two additions: we also consider the 8 open maps from \microrts AI
competitions, and we also run experiments within a chaotic environment
to test  \microphantom 's decision-making robustness.

The baseline bot has again the  same code than \microphantom except for
the  unit production  behavior.  There  is one  domain knowledge-based
action coded into the  unit production behavior, forcing \microphantom
to produce  twice the fastest  unit to produce  at the beginning  of a
game  on  a  small map  (with  a  surface  smaller  than 144,  \ie,  a
$12 \times 12$ grid map).  This domain knowledge-based action has been
disabled during experiments to have a fair comparison of \microphantom
and its baseline on small maps.

\microphantom asks  the constraint solver  to decide about  what units
should be produced at each frame  where at least one barracks is ready
for training. Solving the Unit Production Problem is done within 100ms,
whatever the situation: map size, army composition, etc.

Finally, all maps  have been played 100 times by  each bot, versus the
POLightRush bot.   Our bots played 50  times at the Player  1 starting
position (North-West  on basic maps, but  it can be elsewhere  on open
maps) and  50 times at the  Player 2 starting position  (South-East on
basic maps),  on each map.   In chaotic environments, our  bots played
once  at  the  Player 1  and  the  Player  2  position with  the  same
configuration   setting,  \ie,   the  same   attributes  of   the
game. Then, this  configuration setting is randomized  before each new
couple of games.

\begin{table}[ht!]
  \centering{
    \begin{tabular}{|r||c|c|}
      \hline
      Bots & Basic maps & Open maps\\
      \hline
      Baseline & 61.75 & 64.87\\
      \microphantom & 73.25 & 76.00\\
      \microphantom chaos & 70.91 & 67.93\\
      \hline
    \end{tabular}
  }
  \caption{Score averages of \microphantom and its associated
    baseline on basic and open maps}\label{tab:microphantom_xp}
\end{table}

Table~\ref{tab:microphantom_xp} compiles  4,200 games in  total: 1,800
on basic maps and 2,400 on  open maps. It shows averages of normalized
scores for \microphantom, both in  fixed and chaotic environments, and
its  baseline in  a fixed  environment.   Notice that  the version  of
POLightRush used  for games  in Table~\ref{tab:microphantom_xp}  is an
enhanced   version   compare   to   the  one   used   for   games   in
Table~\ref{tab:poadaptive_xp}.

We can see  that, despite playing against a  stronger POLightRush bot,
results    of    the    baseline    based    on    \microphantom    in
Table~\ref{tab:microphantom_xp} are  greatly better than  results from
the  baseline based  on \poadaptive  in Table~\ref{tab:poadaptive_xp}.
This difference is  due to domain knowledge-based actions  we added in
\microphantom. We are then able  to quantify improvements due to these
hard-coded modifications:  they lead  to an increase  of approximately
50\% of  the win rate  against POLightRush  bot.  A finer  analysis of
result data tells  us that this gain comes mostly  from the conversion
of draw games into won games.

Table~\ref{tab:microphantom_xp}  shows  a significant  improvement  of
\microphantom compare  to its baseline, clearly  more significant than
\poadaptive results in Table~\ref{tab:poadaptive_xp}. This is only due
to our improved unit production behavior compare to \poadaptive.

\microphantom 's results on chaotic  environments are similar to fixed
ones,  showing that  \microphantom is  perfectly able  to handle  rule
changes without  disturbing its decision-making process.  The score on
open maps seems significantly different  though, with a score of 67.93
in chaotic  environments against 76  in fixed ones. Actually,  this is
only  due  to  one  open   map,  NoWhereToRun,  which  is  very  small
($9 \times 8$) where  the two players are separated by  a thin wall of
resources.   Usually,   \microphantom   nearly  always   win   against
POLightRush bot on this map, but in chaotic environments, workers have
one chance over two  to be able to carry 2 resources  instead of 1. In
that configuration,  a hole is very  quickly made in the  wall and let
enter  light units  from  the POLightRush  bot. A  close  look at  the
results data show us that \microphantom  is nearly winning 50\% of the
time  against POLightRush  bot on  this map  in a  chaotic environment
(45~wins, 7~ties  and 48~losses).   Actually, \microphantom  wins when
workers can only carry one resource  at the time, letting it more time
to  prepare  its  defenses,  and   loses  when  workers  can  carry  2
resources.  Without this  very  specific  situation, \microphantom  's
score would be  similar in both fixed and chaotic  environments: if we
consider the same score than the  fixed one on NoWhereToRun, the score
of \microphantom in chaotic environments on open maps turns to be 74.

\microphantom is also  theoretically able to handle rule  changes in a
middle of  a game (a truly  chaotic environment), but this  seems not
easy to process with the current version of \microrts.

\section{Conclusion and perspectives}\label{sec:conclusion}

In this paper, we present  our bot \microphantom playing \microrts. We
show its main differences and  improvements compare to its predecessor
\poadaptive, winner of the partially  observable track of the 2018 and
2019  \microrts  AI  competition,  and show  through  an  experimental
evaluation  including   4.200  games   that  we   achieve  significant
improvements thanks to better decision-making under uncertainty.

We  also  make   \microphantom  able  to  handle   rule  changes:  the
decision-making mechanism in the bot is resilient to the modification
of many game attributes.

Unfortunately, the 2020  edition of the \microrts  AI competition does
not  propose  a  non-deterministic   track  anymore,  unlike  previous
editions. This track run games  where each attack makes damage between
a range fixed for each unit type.  We think such a track has its place
in  the  AI competition  since  it  limits hard-coded,  scripted  bots
tailored for the  competition and force them to  develop and implement
AI techniques to get around non-determinism,  but we guess it has been
removed due to a lack of participants.  We propose that next \microrts
AI competition should contain an  even bolder chaotic track where many
if not all game attributes change  at each game, or even during a
game.

We may improve \microphantom  's \cp model by defining better
error  functions. This  could be  done for
instance  by  using  the  automatic method  proposed  by  Richoux  and
Baffier~\cite{RB20}  to   find  good   error  functions.

Finally,  \microrts  is  certainly  too minimalist  to  contains  many
challenging  combinatorial  optimization  problems,  with  or  without
uncertainty. We  plan to develop  a StarCraft bot using  massively the
decision-making  method presented  in this  paper to  tackle different
aspects of the game, from  economic development and strategy decisions
to micro-management.

\bibliographystyle{IEEEtranS}
\bibliography{microPhantom}

\begin{thebibliography}{10}
\providecommand{\url}[1]{#1}
\csname url@samestyle\endcsname
\providecommand{\newblock}{\relax}
\providecommand{\bibinfo}[2]{#2}
\providecommand{\BIBentrySTDinterwordspacing}{\spaceskip=0pt\relax}
\providecommand{\BIBentryALTinterwordstretchfactor}{4}
\providecommand{\BIBentryALTinterwordspacing}{\spaceskip=\fontdimen2\font plus
\BIBentryALTinterwordstretchfactor\fontdimen3\font minus
  \fontdimen4\font\relax}
\providecommand{\BIBforeignlanguage}[2]{{%
\expandafter\ifx\csname l@#1\endcsname\relax
\typeout{** WARNING: IEEEtranS.bst: No hyphenation pattern has been}%
\typeout{** loaded for the language `#1'. Using the pattern for}%
\typeout{** the default language instead.}%
\else
\language=\csname l@#1\endcsname
\fi
#2}}
\providecommand{\BIBdecl}{\relax}
\BIBdecl

\bibitem{AR19}
V.~Antuori and F.~Richoux, ``{Constrained optimization under uncertainty for
  decision-making problems: Application to Real-Time Strategy games},'' in
  \emph{Proceeding of the Congress on Evolutionary Computation (CEC)}.\hskip
  1em plus 0.5em minus 0.4em\relax IEEE, 2019, pp. 450--457.

\bibitem{Hnich2011}
B.~Hnich, R.~Rossi, S.~A. Tarim, and S.~Prestwich, \emph{{A Survey on CP-AI-OR
  Hybrids for Decision Making Under Uncertainty}}.\hskip 1em plus 0.5em minus
  0.4em\relax Springer New York, 2011, pp. 227--270.

\bibitem{microRTS}
S.~Ontañón, ``{The Combinatorial Multi-armed Bandit Problem and Its
  Application to Real-time Strategy Games},'' in \emph{Proceedings of the 9th
  AAAI Conference on Artificial Intelligence and Interactive Digital
  Entertainment ({AIIDE'13})}, 2014, pp. 58--64.

\bibitem{microRTScompetition}
S.~Ontañón, N.~A. Barriga, C.~R. Silva, R.~O. Moraes, and L.~H.~S. Lelis,
  ``The first microrts artificial intelligence competition,'' \emph{AI
  Magazine}, vol.~39, no.~1, pp. 75--83, 2018.

\bibitem{RDU_quiggin_82}
J.~Quiggin, ``{A Theory of Anticipated Utility},'' \emph{Journal of Economic
  Behavior \& Organization}, vol.~3, pp. 323--343, 1982.

\bibitem{quiggin1993generalized}
------, \emph{{Generalized Expected Utility Theory: The Rank Dependent
  Model}}.\hskip 1em plus 0.5em minus 0.4em\relax Springer, 1993.

\bibitem{RB20}
F.~Richoux and J.-F. Baffier, ``{Automatic Cost Function Learning with
  Interpretable Compositional Networks},'' \emph{arXiv}, 2020.

\bibitem{RUB16}
F.~Richoux, A.~Uriarte, and J.-F. Baffier, ``{GHOST: A Combinatorial
  Optimization Framework for Real-Time Problems},'' \emph{IEEE Transactions on
  Computational Intelligence and AI in Games}, vol.~8, no.~4, pp. 377--388,
  2016.

\bibitem{VJ05}
G.~Verfaillie and N.~Jussien, ``{Constraint solving in uncertain and dynamic
  environments: A survey},'' \emph{Constraints}, vol.~10, no.~3, pp. 253--281,
  2005.

\bibitem{SCSP}
T.~Walsh, ``{Stochastic Constraint Programming},'' in \emph{Proceedings of the
  15th Eureopean Conference on Artificial Intelligence ({ECAI'02})}, 2002, pp.
  111--115.

\end{thebibliography}

\end{document}